\documentclass[sn-mathphys-num]{sn-jnl}

\usepackage{graphicx}%
\usepackage{multirow}%
\usepackage{amsmath,amssymb,amsfonts}%
\usepackage{amsthm}%
\usepackage{mathrsfs}%
\usepackage[title]{appendix}%
\usepackage{xcolor}%
\usepackage{textcomp}%
\usepackage{manyfoot}%
\usepackage{booktabs}%
\usepackage{algorithm}%
\usepackage{algorithmicx}%
\usepackage{algpseudocode}%
\usepackage{listings}%
\usepackage[utf8]{inputenc}
\newcommand{\MPE}{\textsc{MedPromptExtract}}
\begin{document}
\title[Article Title]{MedPromptExtract (Medical Data Extraction Tool):
Anonymization and Hi-fidelity Automated data extraction using
NLP and prompt engineering}

\author[1]{\fnm{Roomani} \sur{Srivastava}}\email{roomani@iitb.ac.in} 
\equalcont{These authors contributed equally to this work.}

\author[1]{\fnm{Suraj} \sur{Prasad}}\email{surajprasad8977@gmail.com}
\equalcont{These authors contributed equally to this work.}

\author[1]{\fnm{Lipika} \sur{Bhat}}\email{lipika.u@gmail.com}

\author[2]{\fnm{Sarvesh} \sur{Deshpande}}\email{sarvesh.ssd7@gmail.com}

\author[3]{\fnm{Barnali} \sur{Das}}\email{drbarnalid@gmail.com}

\author*[1]{\fnm{Kshitij} \sur{Jadhav}}\email{kshitij.jadhav@iitb.ac.in}
\affil[1]{\orgname{Indian Institute of Technology, Bombay}, \orgaddress{\street{IIT Area, Powai}, \city{Mumbai}, \postcode{400076}, \state{Maharashtra}, \country{India}}}

\affil[2]{\orgname{Sardar Patel Institute of Technology}, \orgaddress{\street{Bhavan's Campus, Munshi Nagar, Andheri West}, \city{Mumbai}, \postcode{400058}, \state{Maharashtra}, \country{India}}}

\affil[3]{\orgdiv{Department}, \orgname{Kokilaben Dhirubhai Ambani Hospital and Medical Research Institute}, \orgaddress{\street{Rao Saheb Achutrao Patwardhan Marg, Four Bungalows, Andheri West}, \city{Mumbai}, \postcode{400053}, \state{Maharashtra}, \country{India}}}

\abstract{\textbf{Introduction:} The labour-intensive nature of data extraction from sources like discharge summaries (DS) poses significant obstacles to the digitisation of medical records particularly for low- and middle-income countries (LMICs). In this paper we present a completely automated method \textbf{\MPE} to efficiently extract data from DS while maintaining confidentiality. 

\textbf{Methods:} The source of data was Discharge Summaries (DS) from Kokilaben Dhirubhai Ambani Hospital (KDAH) of patients having Acute Kidney Injury (AKI). A pre-existing tool EIGEN which leverages semi-supervised learning techniques for high-fidelity information extraction was used to anonymize the DS, Natural Language Processing (NLP) was used to extract data from regular fields. We used Prompt Engineering and Large Language Model(LLM) to extract custom clinical information from free flowing text describing the patient's stay in the hospital. Twelve features associated with occurrence of AKI were extracted. The LLM's responses were validated against clinicians annotations. 
\par \textbf{Results:} The \MPE tool first subjected DS to the anonymization pipeline which took three seconds per summary. Successful anonymization was verified by clinicians, thereafter NLP pipeline extracted structured text from the anonymized pdfs at the rate of 0.2 seconds per summary with 100\% accuracy. Finally DS were analysed by the LLM pipeline using Gemini Pro for the twelve features. Accuracy metrics were calculated by comparing model responses to clinicians annotations with seven features achieving AUCs above 0.9, indicating high fidelity of the extraction process.
\par \textbf{Conclusion:} \textbf{\MPE } serves as an automated adaptable tool for efficient data extraction from medical records with a dynamic user interface.}

\keywords{Digitizing Medical Records, Automated Anonymisation, Information Retrieval, Large Language Models, Prompt Engineering}
\maketitle

\section{Introduction}\label{sec1}

Extracting relevant data from non digitized medical records can be a challenging task. 
Additionally, data may be required in retrievable formats for research purposes\cite{lehne2019digital} as well as other downstream applications. Discharge summaries (DS), which provide comprehensive information of a patient's history, treatment, and post-discharge care instructions, are valuable sources of such data \cite{komenan2023qualitative}. Anonymizing such records is essential to maintain confidentiality when used for downstream applications \cite{rothstein2010deidentification}. Moreover, the implementation of artificial intelligence (AI) and machine learning (ML), along with the adoption of paperless workflows, is crucial for transforming healthcare data and improving clinical decision-making. 

A comprehensive review by Meystre et al \cite{meystre2010automatic} highlights common methods used to automate removal of protected health information (PHI) from medical text documents which are observed to be either pattern recognition based or ML based. Pattern recognition uses rules, dictionaries and knowledge resources, while ML requires large sets of annotated records to effectively identify and remove such information. 
For example, work performed by Deleger et al \cite{deleger2013large} achieved de-identification of clinical records of 22 different medical record formats with high precision and recall scores utilizing  MITRE Identification Scrubber Toolkit \cite{aberdeen2010mitre} which works on the procedure of \textit{tag-a-little, learn-a-little} by annotating small sets of EHRs, training a model followed by hand correcting it which is labour intensive. More recent work by Mao et al \cite{mao2019hadoken} which utilizes BERT-CRF model for token level classification for anonymisation of medical records also requires fine tuning BERT on a large corpus. 

Most of the work for data extraction from structured text is focused on extracting data directly from Electronic Medical Records (EMRs). Recent work in this domain utilize Natural Language Processes (NLP) such as use of regular expressions to extract data \cite{c2021automated}, whereas the older works mostly depend on utilizing Structured Query Language (SQL) \cite{kristianson2009data}. Health data management in LMICs such as India run on a hybrid system where most records are maintained offline such as DSs with only partial records are available in the digitized form resulting in limited information available through EMRs. Extraction of information related to medication from DS have been summarised in Doan et al \cite{doan2012recognition} and that for other clinical entities such as medical problems, tests, and treatments have also been explored by Jiang et al \cite{jiang2011study}. Both these works implement Name Entity Recognition (NER) using ensemble classifiers which requires annotated datasets which is a labour intensive process. 

Data extraction from unstructured data has been attempted by several researchers even prior to the era of Large Language Models (LLM), where NER has been extensively used. Most recently, work performed by Adamson et al \cite{adamson2023approach} extracts information in terms of binaries (yes/no, present/absent) or allocate the patient document to a particular category (malignant/non-malignant). In their process they first utilize NLP techniques to identify phrases of interest for extraction, where NLP derived features are inputs to a model which aims to assign the patient document to a particular class within a categorical variable. Similar efforts were made by Hong et al \cite{hong2020annotation} and Topaz et al \cite{topaz2016automated} where they used rule based system and pre-validated dictionary based NLP pipelines respectively to extract information from medical texts. However, they do acknowledge that such systems may fail where complex information is considered such as description of medical events. In the same spectrum work by Lee et al \cite{lee2018conditional} demonstrate that conditional random field (CRF) are more efficient for the purpose of data extraction from unstructured text. 
A couple of research papers focusing on use of LLMs for medical data extraction have also been reported. One of which focused on using LLMs to accelerate annotating medical records for efficient extraction \cite{goel2023llms} and second carried out the extraction process for substance abuse disorders albeit requiring some post processing steps to refine the outputs\cite{mahbub2024leveraging}. 

\par In this paper, we present \textbf{\MPE} a fully automated tool for effectively extracting data from discharge summaries while ensuring confidentiality. We first anonymise the records using EIGEN (Expert-Informed Joint Learning aGgrEatioN)\cite{singh2023eigen}, a name-entity-recognition framework. On the anonymised records we implement Natural Language Processing, specifically the RegEx algorithm\cite{locascio2016neural} for detection of regular expressions such as "Chief Complaint" or "Discharge date". While portions of the discharge summary with single word or expression field could be reproduced very well with RegEx, the larger task remained that of extracting relevant information from the "Course in Hospital" section which details patients' stay in the hospital as described by the doctors. 
Here, we use prompt engineering \cite{chen2023unleashing} with LLMs to determine occurrence of events of relevance from the free flowing text in the documents. 
In our work we demonstrate the utility of LLMs to extract context aware relevant information even without fine tuning (zero-shot) on a corpus specific to this research problem  without having to annotate the clinical text beforehand.

\section{Methodology}\label{sec2}
\subsection{Dataset}
This study was a part of our long term goal to identify early predictors of Acute Kidney
Injury (AKI) in patients admitted to Kokilaben Dhirubhai Ambani Hospital (KDAH),
Mumbai, India. Ethical clearance was obtained from the Institutional Ethical Committee (IEC) of KDAH, Mumbai (IEC - A Code: 037/2021). Serum creatinine values, a biochemical product which is marker for AKI as per KDIGO 2012\cite{khwaja2012kdigo} and AACC\cite{el2021aacc} 2020 guidelines, were collected for patients admitted to the hospital between 1st January to 31st August 2023. The serum creatinine readings were extracted from the Laboratory Information System (LIS) using Structured Query Language. Patients with a single serum creatinine reading, outpatient department (OPD) patients and patients with length of stay less than 24 hours were excluded from our study. In the event that a patient had numerous admissions in the study period, only one encounter was chosen while remaining encounters were excluded.
The KDIGO 2012\cite{khwaja2012kdigo} was employed to label the serum creatinine readings as AKI, by medical domain experts. The baseline creatinine was determined by measuring the lowest serum creatinine reading during the hospital admission \cite{graversen2022defining}. DS of the admitted patients was collected from the Hospital Information System (HIS): Dedalus, iSoft 12.14.1.  A total of 914 DS were retrospectively collected from the patients who had AKI based on serum creatinine values. All DS followed a standard format, facilitating automated data extraction.

\subsection{Anonymisation of discharge summaries}
We used \textbf{EIGEN} \cite{singh2023eigen} for anonymization, leveraging semi-supervised learning techniques for high-fidelity information extraction from documents.
Firstly Document Text Recognition (DocTR) \cite{doctr2021} was used to convert PDF files to JSON files for compatibility with EIGEN. This model employs optical character recognition (OCR) and layout analysis to extract text and spatial structure from the DS. Labeling functions (LFs),which are functions or heuristics that generate noisy or approximate labels for training data \cite{ratner2016data}, were designed using positional heuristics to generate surrogate labels. These surrogate labels serve as a form of weak supervision, enabling the model to learn from both a smaller set of high-confidence, human-annotated data and a larger set of data labeled through these heuristics. This approach helps to extend the training dataset effectively without the need for extensive manual annotation, significantly reducing annotation costs. Our training set consisted of 714 documents, of which only 20 were labeled, minimizing annotation costs.
Following the generation of LFs, the model then has to aggregate labels from these LFs to assign specific class labels to the unlabelled data in the training set. These class labels will identify data which has to be masked and data which has to be extracted. In our work, we had four classes to label: name, patient ID, location, and others (which represents all the information we wish to extract).  
This is achieved by a joint fine-tuning method of two models which may be referred to as the "feature model" and the probabilistic or "graphical model".
The graphical model in the Eigen framework is a probabilistic generative model which assumes independence between LFs and learns their reliability scores through a set of parameters corresponding to each class. The graphical model establishes a joint distribution between LF outputs and the true labels, reconciling conflicts when LFs assign different labels to the same word. This approach is combined with a deep learning feature-based model for which EIGEN uses LayoutLM \cite{10.1145/3394486.3403172}, a pre-trained deep neural network, which predicts the probability distribution of class labels based on word features and spatial information. The overall system utilizes joint learning, where both the graphical model and the feature model are trained together using labeled and unlabeled data. A Kullback-Leibler (KL) divergence term is included in the joint learning to ensure that both models (the graphical model and the feature-based model) agree on the predictions across labeled and unlabeled data. 
\par The model was trained with a batch size of 32. The  learning rate for the feature model and graphical model was $5 \times 10^{-5}$ and 0.01 respectively. Data was trained on 4X48 GB NVIDIA A6000 GPU Server. The optimizer used was AdamW. This architecture enables efficient semi-supervised learning, enhancing label consistency and improving the performance of information extraction. 

\begin{figure}[h] 
\centering
\includegraphics[width=0.9\textwidth]{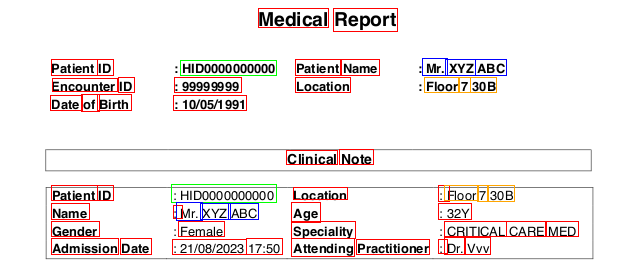}  
\caption{Visualizing output of EIGEN representing four class labels: "name" (Blue), "patient ID" (Green), "location" (Yellow), and "others"(Red). The class "others" represents all the information to be extracted while the remaining classes were masked. }
\label{fig:1}
\end{figure}

We used predicted class labels to anonymize DS by removing patient names and locations, replacing patient IDs with serial numbers (Algorithm \ref{alg:1}). This process was conducted within the medical records department of hospital, ensuring complete confidentiality. Figure \ref{fig:1} shows the visualization of bounding boxes for class labels after implementing EIGEN.

\begin{algorithm}[h]
\caption{Anonymisation of Discharge Summaries}
\label{alg:1}
\begin{algorithmic} [1]
\State \textbf{Getting surrogate labels}: Define text triggers for LFs, define positional heuristics
    \For{Each Labelling Function}
        \State Match text triggers with words in the document
        \If{Text matched and bounding box satisfies positional heuristics}
            \State Assign class label to the word
        \Else
            \State Abstain
        \EndIf
    \EndFor
    \State Load data - Discharge Summaries, Set optimizer to AdamW
    \For{Epoch in range of number of epochs}
        \State Train feature model (LLM), graphical model
        \State \textbf{Output labels for all words for each document in the dataset}
        \State Calculate Total Loss
        \State Back propagate losses and update model weights
    \EndFor

\Return \textbf{Predicted Class Labels}
\end{algorithmic}
\end{algorithm}

\subsubsection{NLP Model Architecture}
A standard discharge summary is semi structured where regular headings like "Chief complaint" or "Diagnosis" are followed by the text corresponding to them. This structure of repetitive terms, was first utilized to extract all text from the DS in an organized format of a DataFrame. 
\begin{itemize}
\item \textbf{Use of Regular Expressions}:
The term "regular expressions" in NLP refers to a sequence of characters or notations to represent a set of strings. In our study we employed such regular expressions to match specific patterns within the text, facilitating information extraction which follow these string representations. We termed these string representations as 'stopwords'.
\item \textbf{Stopword Filtering}:
Certain terms which represented patient identification phrases such as "patient id" or "location" were prevented from being extracted by employing a 'stopword filtering' mechanism which ensured exclusion of irrelevant text sections.
\end{itemize}
The extracted information was organized and stored in a structured DataFrame, with each stopword serving as a feature label for information following it (Algorithm \ref{alg:2}). Data was saved under feature labels such as- date of admission, discharge, diagnosis, chief complaints, past history, significant findings and investigations, medications administered during the stay and those recommended on discharge.  The unstructured data under the heading “Course in Hospital” had all the physicians notes as free flowing text which was processed using the third aspect of our pipeline explained in the next section.

\begin{algorithm}[h]
\caption{Extract Patient Information from anonymised Discharge Summaries}
\label{alg:2}
\small
\begin{algorithmic}[1]
    \Require Folder containing PDF files
    \Ensure DataFrame containing extracted information
    \State Define \texttt{stop\_keywords} as a list of keywords to halt text extraction
    \State Set \texttt{folder\_path} to the directory containing the PDF files
    \State Initialize an empty DataFrame \texttt{df} with specified columns
        \For{each filename in the \texttt{folder\_path}}
            \If{the filename ends with ".pdf"}
            \State Construct the full file path
            \State Initialize a dictionary \texttt{data\_dict} to store extracted data
            \State Initialize an empty string \texttt{extracted\_text}
            \For{each page in the PDF}
                \State Extract text from the page, Append the extracted text to \texttt{extracted\_text}
            \EndFor
            \State Replace newline characters in \texttt{extracted\_text} with spaces
            \State Use regular expressions to extract data based on predefined patterns and \texttt{stop\_keywords}
            \State Store extracted information in \texttt{data\_dict}
            \State Normalize \texttt{data\_dict} to ensure all lists have the same length
            \State Append data from \texttt{data\_dict} to the DataFrame \texttt{df}
            \State Print any exceptions encountered
    \EndIf
\EndFor

\Return \texttt{df} 
\end{algorithmic}
\end{algorithm}

\subsection{Extracting unstructured text with Prompt engineering}
In this step, we analyzed free-flowing text under the heading "Course in Hospital" to identify features associated with occurrence of Acute Kidney injury (AKI). Twelve features were identified through domain knowledge and interactions with nephrologists for extraction using the generative ability of LLMs and prompt engineering. These 12 features were (1) mention of patient having AKI, (2) having undergone angiography, (3) nephrologist consulted or not, (4) diuretic administered, (5) fluid restriction advised, (6) general anesthesia administered, (7) known case of hypertension, (8) admitted to the ICU, (9) undergone imaging procedure with contrast dye being administered, (10) drop in oxygen saturation, (11) patient with tachycardia and (12) if patient required a ventilator.
A single prompt was used to extract all clinical impressions, utilizing the Gemini API\cite{Pichai_2023}. The prompt query was as follows:
"I want you to act like a doctor. I will give you summary of a patient's stay in the hospital you will evaluate it and answer a set of questions as yes or no.
1. Is there any mention of consultation by nephrologist for this patient; 2. Is it mentioned that the patient has Acute Kidney Injury (AKI); 3. Was the patient put under General Anaesthesia at any point; 4. Has hypertension been mentioned as a previously existing condition in the patient; 5. Has the patient been advised to reduce fluid intake; 6. Has this patient undergone angiography; 7. Is the patient being given any diuretic; 8. Has the patient undergone any imaging procedure using contrast dye; 9. As per this summary was the patient ever admitted to the ICU; 10. Was the patient put on Ventilator during his/her stay in the hospital; 11. Did the patient develop Tachycardia at any point during his/her stay in the hospital; 12. Is there any mention of drop in Oxygen saturation, Only return the answers, next to indices number and not the questions."
Gemini pro version 0.1.0, was configured with temperature parameter set to 0 to increase the probability of precise and deterministic outputs. 
Two iterations of model responses were analyzed using Kappa coefficient for intra-model response agreement.

\subsection{Validation with human annotations}
\begin{figure*}[htbp] 
    \centering
    \includegraphics[width=\textwidth, height= 5cm]{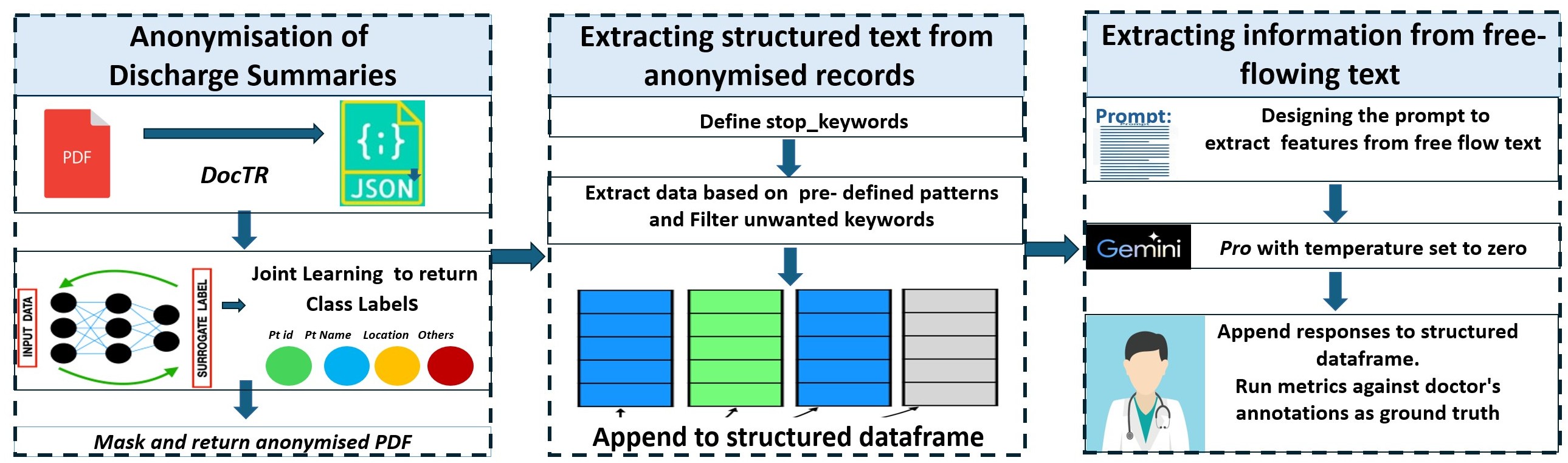}  
    \caption{Workflow of MedPromptExtract Tool: representing the process of anonymizing DS using EIGEN, followed by extracting structured text using NLP and extracting information from free-flowing text using LLM}
    \label{fig:2}
\end{figure*}

 Two clinicians annotated 48 DS, and inter-rater agreement was assessed using the Kappa coefficient. Data extracted using LLM was validated against this as ground truth by calculating accuracy, precision, recall, F1 score and AUC using the ScikitLearn library\cite{scikit-learn} in Python. 
Figure \ref{fig:2} illustrates the pipeline of MedPromptExtract.

\section{Results}
The tool \textbf{\MPE} has three components to its pipeline where we have first anonymized DS using EIGEN followed by data extraction of both structured text and information from free flowing text. Human verification of outputs after each step was done and time efficiency of the tool was assessed.
The anonymization pipeline took approximately three seconds to process and mask one DS, so in total 914 DS were processed in 46 minutes. The outputs of anonymised pdfs were verified by human intervention. It was noted that out of the 914 DS processed, four were not processed correctly   due to format errors in the source PDFs received from the hospital. These 4 pdfs were excluded from any further analysis bringing down the number of DS analysed to 910.
Following anonymization the NLP pipeline utilized RegEx to extract structured text from the DS. The time required to process a single DS was 0.2 seconds, that is 910 pds were fully processed within 2.5 minutes. All data extracted from the NLP pipeline was assessed for completeness and it was noted that all sections from all DS were successfully extracted with 100\% accuracy. However, it was noted that 58 summaries had an empty "Course in Hospital" field, this was verified by checking the original discharge summary. This content was essential from the execution of the next step in the pipeline hence these 58 entries were excluded from further analysis bringing the final sample size to 852.
The next step was to use LLM - Gemini Pro to extract custom information on the 12 features associated with occurrence of AKI (listed earlier) from free flowing text in the section "Course in Hospital" of 852 discharge summaries. Fifty of the DS were randomly selected for annotations by two clinicians for these 12 features. Inter rater agreement (Kappa) between the two annotators was also computed and found to be more than 0.8 for all 12 features. Two iterations of the LLM were run to make sure that the responses were deterministic enough. This was also verified by applying Kappa coefficient for intra-model agreement, this was found to be 1.

The compute time required for the LLM to process one DS was nearly 7 seconds, equating to about 100 minutes for 852 DS. On the other hand time taken for Human Annotations was 9.6 man-hours for 48 annotations at the rate of 12 minutes per document. The intention here is not to compare man and machine, where machine is certainly more time efficient, but to emphasise the precious time spent by clinician on such tasks which can be invested elsewhere.
Finally the models responses were checked against clinician's annotations which was considered the ground truth. Any conflict in the agreement of the two clinicians was resolved by discussion and a single response was noted for each of the 12 features being evaluated. The response of the model was considered the predicted label and metrics were calculated against the ground truth. Accuracy, Sensitivity, Specificity, Precision, Recall and Area Under the Curve (ROC-AUC) were calculated as detailed in table \ref{tab:1}. Seven of the twelve features has AUCs above 0.9 indicating high fidelity of the data extraction process. These features were - patient undergone angiography; diuretic given or not; fluid restriction advised; hypertension; imaging procedure with contrast; patient having tachycardia and use of ventilator. Furthermore another four features had AUCs above 0.7. Poorest AUC was noted for the feature "Consultation by nephrologist". The possible reasons for poor metrics of some features are discussed in the next section.

\begin{table}[]
\caption{Validation of model responses with clinicians' responses}
\label{tab:1}
\centering
\begin{tabular}{@{}lcccccc@{}}
\toprule
\textbf{Feature} &
  \multicolumn{1}{l}{\textbf{Acc.}} &
  \multicolumn{1}{l}{\textbf{Sen.}} &
  \multicolumn{1}{l}{\textbf{Spec.}} &
  \multicolumn{1}{l}{\textbf{Pre.}} &
  \multicolumn{1}{l}{\textbf{F1}} &
  \multicolumn{1}{l}{\textbf{AUC}} \\ \midrule
AKI Mentioned                   & 0.96 & 0.5  & 1    & 1    & 0.67 & 0.75 \\
Angiography Done                & 0.98 & 1    & 0.98 & 0.86 & 0.92 & 0.98 \\
Consultation by Nephrologist    & 0.81 & 0.12 & 0.95 & 0.33 & 0.18 & 0.53 \\
Diuretic Given                  & 0.98 & 1    & 0.97 & 0.92 & 0.96 & 0.98 \\
Fluid Restriction Advised       & 0.98 & 1    & 0.98 & 0.67 & 0.8  & 0.98 \\
General Anaesthesia             & 0.81 & 0.65 & 0.9  & 0.79 & 0.71 & 0.77 \\
Hypertension                    & 0.98 & 1    & 0.98 & 0.75 & 0.86 & 0.98 \\
ICU Admission                   & 0.71 & 0.5  & 0.95 & 0.93 & 0.65 & 0.72 \\
Imaging Procedure with Contrast & 0.94 & 0.88 & 0.95 & 0.78 & 0.83 & 0.9  \\
Oxygen Saturation Drop          & 0.96 & 0.5  & 1    & 1    & 0.67 & 0.75 \\
Tachycardia                     & 1    & 1    & 1    & 1    & 1    & 1    \\
Ventilator Used                 & 0.98 & 1    & 0.98 & 0.8  & 0.89 & 0.98 \\ \midrule
\multicolumn{7}{l}{\begin{tabular}[c]{@{}l@{}}\textit{Note: Acc. = Accuracy, Sen. = Sensitivity, Spec. = Specificity,} \\ \textit{Pre. = Precision, F1 = F1 Score, AUC = Area Under the Curve}\end{tabular}} \\ \bottomrule
\end{tabular}
\end{table}

\section{Discussion and Limitations}
In contemporary medical practice, data extraction is hindered by several factors, including the lack of information system integration, reliance on manual workflows, excessive workloads, and the absence of standardized databases. These combined limitations often result in clinically significant data being overlooked, which can negatively impact clinical outcomes. This includes the potential for medical errors, reduced diagnostic confidence, overuse of medical services, and delays in diagnosis and treatment planning. To address this, several tools have been developed to extract and analyse medical data from electronic health records.
One such tool is HEDEA, a Python-based system that extracts structured information from various medical records using regular expressions, focusing on the Indian healthcare context \cite{aggarwal2018hedea}. The tool specifically targets the challenges posed by the semi-structured nature of medical data. It also solely relies on regular expression while it can incorporate additional sophisticated tools to increase the analytical power. The study does not explicitly report precision and accuracy, which would be crucial in evaluating the performance of the tool. Another tool, MetaMed, is an open-source application, that extracts metadata from heterogeneous medical data and stores it in a semantically interoperable format \cite{vcelak2012metamed}. Data extraction using MetaMed could be challenging due to the complexity in the source of data. Additionally, the extracted data is not compared with ground truth, making it difficult to evaluate the accuracy of the tool. Another tool was developed by Liljeqvist et al. for extracting information from influenza-like illness (ILI) by searching both in free text and in coded data \cite{liljeqvist2011automated}. Two public health physicians examined the electronic medical records of two primary health clinic over the course of a week in order to assess the sensitivity and specificity of retrieving pertinent ILI cases by free text search. There was no examination of the extracted material's consistency with its original source. All the tools lack an automated anonymization step to ensure data privacy. A comprehensive end-to-end tool capable of both anonymizing discharge summaries and extracting pertinent information from medical health records is currently unavailable. Additionally, direct comparison between extraction tools poses challenges since each EHR provider requires the development of a new extraction tool relevant to their sources of input data.

In our approach we demonstrate high fidelity data extraction in response to custom prompts without any fine tuning of the model used or annotations performed on the input text. Seven out of twelve features in our tested prompt achieve area under the curve of more than 0.9, the remaining five did not perform so well; the reasons for this are discussed here. 
\par To investigate this, randomly selected DS giving positive and negative responses to these five questions were checked by experts to qualitatively assess its responses. It was noted that the LLM was delivering contextual responses to certain questions. For example, for the feature determining if \textbf{AKI MENTIONED OR NOT}, a reference to rise in creatinine values in the text also resulted in the LLM giving a positive response. This may not have been correct in every case. Similarly, in questions such as \textbf{WHETHER GENERAL ANAESTHESIA WAS GIVEN} or \textbf{WHETHER PATIENT WAS ADMITTED TO THE ICU}, responses by the LLM were contextually derived. Another example is, patient having undergone any kind of surgery was assumed to have undergone general anaesthesia by the model, which is not always true. Other questions which suffered from such assumption was \textbf{OXYGEN SATURATION} being equated with any kind of respiratory issues. The question on \textbf{CONSULTATION BY NEPHROLOGIST} was a unique case as the LLM was unable to match human annotations as in most summaries the name of the nephrologist was referred to rather than the speciality, resulting in the doctor annotations identifying it correctly (as they were from the same hospital), but the LLM not being able to match it.
\par 
A work around this problem could be to give more conditional prompts to the LLMs prior to the questions or even few shot examples to identify special scenarios where for instance ICU admission is not necessary after a minor surgery, which would be explored in the future.  
A limitation of \textbf{\MPE} is that our pipeline is custom-developed for the DS of the source hospital. While generalizable, it would require minor adjustments to the training pipeline of EIGEN and the prompts to be used for other similar use-cases.

\section{User Interface}
The user interface (UI) of \textbf{\MPE} has been developed for this end-to-end pipeline and is demonstrated in the video attached. The interface is simple and clear with intuitive navigation options and options for saving/downloading outputs at every point as per the need. The UI also provides the option to edit the prompt to derive information that may be different from what we have demonstrated here. For the purpose of demonstration, in the video, we demonstrate both the options of using our predefined prompt or using a different prompt and visualizing the output of both. We demonstrate processing of a single DS for the purpose of understanding and thereafter we also display the ability of our UI to process bulk summaries with processing 20 at the same time which took less than 5 minutes to process. This system has been deployed in the source hospital. A demonstration video of the UI is available here: \href{https://www.dropbox.com/scl/fi/9qqpj7phjuwd8zkyody6f/MedPromptExtract.mp4?rlkey=suhq569cwm3nk1ug82jbqvjkq&st=bxi7dy8g&dl=0}{MedPromptExtract Demo Video}.

\section{Conclusion}
Implementing ML tools in real life hospital scenarios is a challenge given the practical constraints set by data quality and availability. This work brings forth the ability of our methodology to perform automated anonymisation and data extraction with hi-fidelity utilizing very less labelled data, significantly reducing anonymisation costs, thus providing a framework for digitization and further analysis of healthcare data. This method facilitates making the data available in readily consumable format while maintaining anonymity. 
\textbf{\MPE}  has been deployed in the source hospital and will be integrated with their EHR system for downstream applications.


\end{document}